\documentclass[]{bytedance_seed}
\usepackage{dscout_arxiv}
\microtypesetup{expansion=false}

\usepackage{wrapfig}
\usepackage{graphicx}
\usepackage{multirow}
\usepackage{multicol}
\usepackage{amsthm}
\usepackage{svg}
\usepackage{xcolor}
\usepackage{soul}

\newif\ifrevision
\revisiontrue  

\usepackage{float}
\usepackage{placeins}

\usepackage{wrapfig}
\usepackage{caption}
\usepackage{algorithm}
\usepackage{subcaption}
\usepackage[utf8]{inputenc} 
\usepackage[T1]{fontenc}    
\usepackage{url}            
\usepackage{booktabs}       
\usepackage{tabularx}
\usepackage{amsfonts}       
\usepackage{nicefrac}       
\usepackage{microtype}      
\usepackage{xcolor}         
\usepackage{amsmath}
\usepackage{amssymb}
\usepackage{mathtools}
\usepackage{slashed}
\usepackage{braket}

\usepackage{enumitem}

\usepackage[colorinlistoftodos,textsize=small]{todonotes}

\newif\ifarxivversion
\arxivversiontrue

\providecommand{\mytitle}[1]{\textbf{#1.}\xspace}

\usepackage{graphicx}
\usepackage{tikz}
\usepackage{tikz-cd}
\usepackage{times}
\usepackage{courier}
\usepackage{bm}
\usepackage{physics}
\usepackage{xcolor}
\usepackage{natbib}
\usepackage{mdframed}
\usepackage{nicefrac}
\usepackage{booktabs}
\usepackage{lipsum}
\usepackage{wrapfig,lipsum,booktabs}
\usepackage{blindtext}
\usepackage{algorithm}
\usepackage{algorithmicx}
\usepackage{algpseudocode}
\usepackage{titletoc}
\usepackage{todonotes}
\usepackage{titletoc}
\usepackage{setspace}

\usepackage[capitalize,noabbrev]{cleveref}

\usepackage{CJK}

\usepackage{array}
\usepackage{makecell}

\newcommand{\bea}{\begin{eqnarray}}
\newcommand{\eea}{\end{eqnarray}}

\usepackage{slashed}

\def\({\left(}
\def\){\right)}
\def\[{\left[}
\def\]{\right]}

\usepackage{dashbox}
\usepackage{xcolor}
\ifarxivversion\else
\usepackage{colortbl}
\fi
\definecolor{lightyellow}{rgb}{1.0, 0.95, 0.7}
\definecolor{Blue}{rgb}{0, 0, 0.8}
\definecolor{blue}{rgb}{0,0,1}
\definecolor{darkgreen}{rgb}{0,0.40,0}
\definecolor{firebrick}{rgb}{0.698,0.133,0.133}

\definecolor{colorA}{rgb}{1,0,0}
\definecolor{colorB}{rgb}{0,0.3,1}
\definecolor{colorC}{rgb}{0.9,0.8,0.2}
\definecolor{colorD}{rgb}{0,0.65,0}
\definecolor{lesslightgray}{rgb}{0.5,0.5,0.5}
\definecolor{light-gray}{gray}{0.95}

\let\hat\widehat

\newcommand{\bg}{\mathbf{g}}

\let\cite\citep 

\usepackage{amsthm}
\makeatletter
\def\th@remark{%
  \thm@headfont{\bfseries}%
  \normalfont 
  \thm@preskip\topsep \divide\thm@preskip\tw@
  \thm@postskip\thm@preskip
}
\makeatother
\makeatletter
\@ifpackageloaded{tcolorbox}{}{\usepackage[many]{tcolorbox}}
\makeatother
\theoremstyle{definition}

\tcolorboxenvironment{theorem}{
  breakable,
  colback=black!10,
  colframe=white,%
  width=\dimexpr\linewidth+10pt\relax,%
  enlarge left by=-5pt,%
  enlarge right by=-5pt,%
  boxsep=5pt,%
  boxrule=0pt,
  left=0pt,right=0pt,top=0pt,bottom=0pt,
  sharp corners,
  before skip=\topsep,
  after skip=\topsep
}

\tcolorboxenvironment{lemma}{
  breakable,
  colback=black!10,
  colframe=white,%
  width=\dimexpr\linewidth+10pt\relax,%
  enlarge left by=-5pt,%
  enlarge right by=-5pt,%
  boxsep=5pt,%
  boxrule=0pt,
  left=0pt,right=0pt,top=0pt,bottom=0pt,
  sharp corners,
  before skip=\topsep,
  after skip=\topsep
}

\tcolorboxenvironment{corollary}{
  breakable,
  colback=black!10,
  colframe=white,%
  width=\dimexpr\linewidth+10pt\relax,%
  enlarge left by=-5pt,%
  enlarge right by=-5pt,%
  boxsep=5pt,%
  boxrule=0pt,
  left=0pt,right=0pt,top=0pt,bottom=0pt,
  sharp corners,
  before skip=\topsep,
  after skip=\topsep
}

\theoremstyle{definition}

\tcolorboxenvironment{definition}{
  breakable,
  colback=black!10,
  colframe=white,%
  width=\dimexpr\linewidth+10pt\relax,%
  enlarge left by=-5pt,%
  enlarge right by=-5pt,%
  boxsep=5pt,%
  boxrule=0pt,
  left=0pt,right=0pt,top=0pt,bottom=0pt,
  sharp corners,
  before skip=\topsep,
  after skip=\topsep
}
\theoremstyle{remark}

\tcolorboxenvironment{assumption}{
  breakable,
  colback=black!10,
  colframe=white,%
  width=\dimexpr\linewidth+10pt\relax,%
  enlarge left by=-5pt,%
  enlarge right by=-5pt,%
  boxsep=5pt,%
  boxrule=0pt,
  left=0pt,right=0pt,top=0pt,bottom=0pt,
  sharp corners,
  before skip=\topsep,
  after skip=\topsep
}

\tcolorboxenvironment{problem}{
  breakable,
  colback=black!10,
  colframe=white,%
  width=\dimexpr\linewidth+10pt\relax,%
  enlarge left by=-5pt,%
  enlarge right by=-5pt,%
  boxsep=5pt,%
  boxrule=0pt,
  left=0pt,right=0pt,top=0pt,bottom=0pt,
  sharp corners,
  before skip=\topsep,
  after skip=\topsep
}

\crefname{theorem}{Theorem}{Theorems}
\crefname{proposition}{Proposition}{Propositions}
\crefname{lemma}{Lemma}{Lemmas}
\crefname{corollary}{Corollary}{Corollaries}
\crefname{definition}{Definition}{Definitions}
\crefname{assumption}{Assumption}{Assumptions}
\crefname{remark}{Remark}{Remarks}
\crefname{problem}{Problem}{Problems}
\crefname{property}{Property}{property}

\numberwithin{equation}{section}
\numberwithin{theorem}{section}
\numberwithin{proposition}{section}
\numberwithin{definition}{section}
\numberwithin{lemma}{section}
\numberwithin{assumption}{section}
\numberwithin{remark}{section}

\usepackage{lipsum}

\makeatletter
\let\save@mathaccent\mathaccent
\newcommand*\if@single[3]{%
    \setbox0\hbox{${\mathaccent"0362{#1}}^H$}%
    \setbox2\hbox{${\mathaccent"0362{\kern0pt#1}}^H$}%
    \ifdim\ht0=\ht2 #3\else #2\fi
}
\newcommand*\rel@kern[1]{\kern#1\dimexpr\macc@kerna}
\newcommand*\widebar[1]{\@ifnextchar^{{\wide@bar{#1}{0}}}{\wide@bar{#1}{1}}}
\newcommand*\wide@bar[2]{\if@single{#1}{\wide@bar@{#1}{#2}{1}}{\wide@bar@{#1}{#2}{2}}}
\newcommand*\wide@bar@[3]{%
    \begingroup
    \def\mathaccent##1##2{%
        \let\mathaccent\save@mathaccent
        \if#32 \let\macc@nucleus\first@char \fi
        \setbox\z@\hbox{$\macc@style{\macc@nucleus}_{}$}%
        \setbox\tw@\hbox{$\macc@style{\macc@nucleus}{}_{}$}%
        \dimen@\wd\tw@
        \advance\dimen@-\wd\z@
        \divide\dimen@ 3
        \@tempdima\wd\tw@
        \advance\@tempdima-\scriptspace
        \divide\@tempdima 10
        \advance\dimen@-\@tempdima
        \ifdim\dimen@>\z@ \dimen@0pt\fi
        \rel@kern{0.6}\kern-\dimen@
        \if#31
        \overline{\rel@kern{-0.6}\kern\dimen@\macc@nucleus\rel@kern{0.4}\kern\dimen@}%
        \advance\dimen@0.4\dimexpr\macc@kerna
        \let\final@kern#2%
        \ifdim\dimen@<\z@ \let\final@kern1\fi
        \if\final@kern1 \kern-\dimen@\fi
        \else
        \overline{\rel@kern{-0.6}\kern\dimen@#1}%
        \fi
    }%
    \macc@depth\@ne
    \let\math@bgroup\@empty \let\math@egroup\macc@set@skewchar
    \mathsurround\z@ \frozen@everymath{\mathgroup\macc@group\relax}%
    \macc@set@skewchar\relax
    \let\mathaccentV\macc@nested@a
    \if#31
    \macc@nested@a\relax111{#1}%
    \else
    \def\gobble@till@marker##1\endmarker{}%
    \futurelet\first@char\gobble@till@marker#1\endmarker
    \ifcat\noexpand\first@char A\else
    \def\first@char{}%
    \fi
    \macc@nested@a\relax111{\first@char}%
    \fi
    \endgroup
    }
\makeatother

\makeatletter
\newcommand*{\redefinesymbolwitharg}[1]{%
  \expandafter\let\csname ltx#1\expandafter\endcsname\csname #1\endcsname
  \@namedef{#1}{\@ifnextchar{^}{\@nameuse{#1@}}{\@nameuse{#1@}^{}}}%
  \expandafter\def\csname #1@\endcsname^##1##2{%
     \csname ltx#1\endcsname\ifx!##1!\else^{##1}\fi\mathopen{}\mathclose\bgroup\left(##2\aftergroup\egroup\right)
     }%
}
\makeatother
\redefinesymbolwitharg{sin}
\redefinesymbolwitharg{cos}

\arxivversiontrue

\renewcommand{\mytitle}[1]{\vspace{2pt}\noindent{\bfseries #1.}\xspace}

\usepackage{titletoc}

\newcommand{\sys}{DScout\xspace}

\def\Snospace~{\S{}}

\newcommand{\sref}[2]{\hyperref[#2]{#1 \ref{#2}}}

\title{Factorization-Guided Training and Inference for Diffusion-Drafted Speculative Decoding
}

\author[*1,3]{Jiakun Fan}
\author[*2,3]{Min Xiao}
\author[3]{Qiang Wang}
\author[3]{Neiwen Ling}
\author[3]{Zhichao Liu}
\author[2]{Zhihao Jia}
\author[1]{Dimitrios S. Nikolopoulos}
\author[\dagger3]{Zijun Wei}

\affiliation[1]{Virginia Tech}
\affiliation[2]{Carnegie Mellon University}
\affiliation[3]{ByteDance}

\contribution[*]{Equal contribution}
\contribution[\dagger]{Corresponding author}
\correspondence{Jiakun Fan at \email{jiakunfan@vt.edu}; Min Xiao at \email{minxiao@andrew.cmu.edu}; Zijun Wei at \email{zijun.wei@bytedance.com}}

\abstract{Speculative decoding accelerates large language model inference through
parallel verification of tokens proposed by a drafter.
Diffusion drafters further parallelize proposal generation, but greater
proposal capacity or higher mean accepted tokens (MAT) need not yield
higher throughput.
Because verification accepts only a consecutive prefix, early rejections
can render later correct predictions ineffective, limiting MAT gains,
while even higher MAT may not improve throughput if per-round latency
increases.
We introduce \textbf{DScout}, which uses two complementary factorizations
to guide diffusion drafter training and inference for higher throughput.
For training, DScout factorizes position-wise loss into inclusion, weighting,
and prediction loss. This guides the design and combination of mixed-horizon
training, intra-block unmasking, and acceptance-weighted loss to align
supervision with prefix acceptance and improve MAT.
For inference, DScout factorizes throughput as the ratio of MAT to
per-round latency, guiding target-conditioning depth, drafter depth,
and inference block size selection to identify a high-throughput
configuration for each workload.
Across six benchmarks, DScout achieves \textbf{$3.84\times$} and \textbf{$4.14\times$}
mean speedups over autoregressive decoding on Qwen3-4B and Qwen3-8B,
respectively, improving mean speedup over the state-of-the-art baseline
by \textbf{9.1\%} and \textbf{7.1\%}.
Across models and concurrency settings, our inference configuration policy
provides up to \textbf{9.75\%} additional mean speedup over DScout's fixed configuration.
}

\usepackage{array}

\definecolor{LightCyan}{rgb}{0.8, 0.9, 1}
\newcolumntype{b}{>{\columncolor{LightCyan}\hspace{0pt}}c}
\definecolor{cadetgrey}{rgb}{0.57, 0.64, 0.69}
\newcolumntype{g}{>{\columncolor{cadetgrey}\hspace{0pt}}c}

\begin{document}

\maketitle

\section{Introduction}

\label{sec:intro}

Autoregressive (AR) large language models generate tokens sequentially,
requiring a model forward pass for each new token.
Speculative decoding amortizes this cost by using a lightweight drafter
to propose multiple tokens for parallel verification in a single
target-model pass, while preserving the target model's output
distribution~\citep{leviathan2023speculative,chen2023speculative}.
Diffusion drafters further parallelize proposal generation by predicting
multiple draft tokens simultaneously. For example, DFlash conditions a
lightweight block-diffusion model on target-model features to generate
an entire proposal in a single draft pass~\citep{chen2026dflash}.

However, improving parallel draft predictions does not necessarily
translate into higher throughput. Verification accepts draft tokens
only up to the first rejection, so later correct predictions contribute
to mean accepted tokens (MAT) only if the preceding prefix is accepted.
Training should therefore account for this prefix dependence when
allocating supervision to improve MAT. Higher MAT, in turn, improves
throughput only when the additional accepted tokens compensate for
increased draft--verify latency. These challenges motivate jointly
considering prefix dependence in training and the MAT--latency trade-off
at inference.

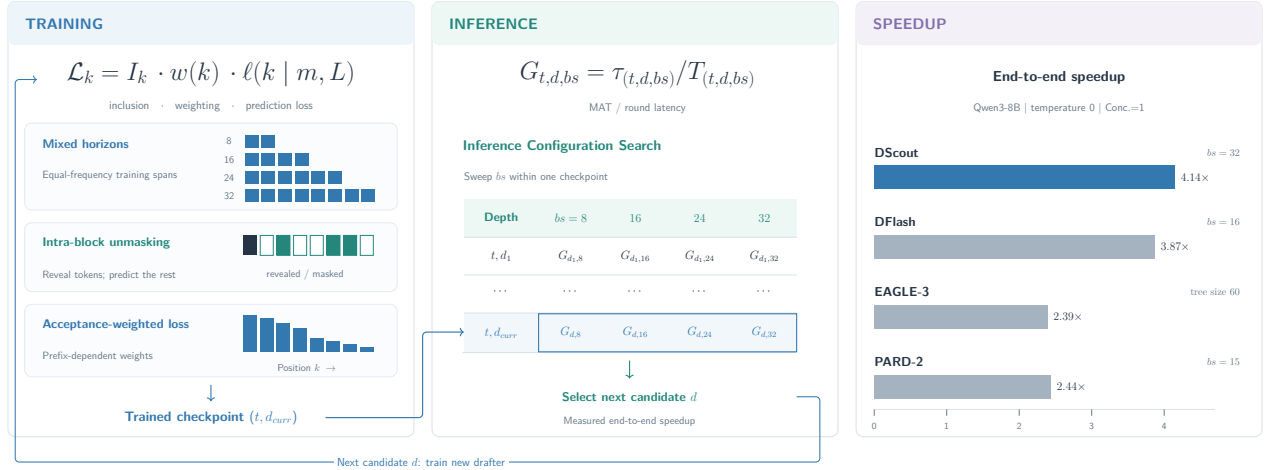
\begin{figure}[!t]
    \centering
     \vspace{-0.2in}
    \resizebox{1\linewidth}{!}{
\begingroup%
\definecolor{softink}{HTML}{263546}%
\definecolor{softmuted}{HTML}{667788}%
\definecolor{softblue}{HTML}{3478B0}%
\definecolor{softteal}{HTML}{25877C}%
\definecolor{softpurple}{HTML}{8470AA}%
\definecolor{softedge}{HTML}{D9E2EA}%
\definecolor{softbar}{HTML}{A5B3C1}%
\definecolor{softbluebg}{HTML}{EDF4FA}%
\definecolor{softgreenbg}{HTML}{EDF7F4}%
\definecolor{softpurplebg}{HTML}{F3F0F8}%
\definecolor{softcard}{HTML}{FAFCFE}%
\begin{tikzpicture}[x=.01in,y=-.0125in,every node/.style={inner sep=0pt,text=softink,font=\sffamily\fontsize{12}{14}\selectfont},line width=.72pt]
\path[use as bounding box] (20,24) rectangle (1750,548);
\foreach \x/\w/\title/\col/\bg in {20/560/TRAINING/softblue/softbluebg,605/560/INFERENCE/softteal/softgreenbg,1190/560/SPEEDUP/softpurple/softpurplebg}{
 \draw[fill=white,draw=softedge,rounded corners=7.2pt] (\x,24) rectangle ++(\w,480);
 \path[fill=\bg,rounded corners=7.2pt] (\x,24) rectangle ++(\w,48);
 \node[anchor=west,text=\col,font=\sffamily\bfseries\fontsize{15.12}{17}\selectfont] at (\x+23,48) {\title};}
\node[font=\fontsize{25.2}{28}\selectfont] at (300,103) {$\mathcal L_k=I_k\,\cdot w(k)\,\cdot\ell(k\mid m,L)$};
\node[text=softmuted,font=\sffamily\fontsize{11.52}{13}\selectfont] at (300,139) {inclusion\quad$\cdot$\quad weighting\quad$\cdot$\quad prediction loss};
\foreach \y/\h in {158/96,268/76,358/80}{\draw[fill=softcard,draw=softedge!70,rounded corners=7.2pt] (43,\y) rectangle ++(514,\h);}
\node[anchor=west,text=softblue,font=\sffamily\bfseries\fontsize{12.5}{15}\selectfont] at (67,180.4) {Mixed horizons};
\node[anchor=west,text=softmuted,font=\sffamily\fontsize{10.5}{15}\selectfont] at (67,215.6) {Equal-frequency training spans};
\foreach \b/\n/\yy in {8/2/171,16/4/191,24/6/211,32/8/231}{
 \node[text=softmuted,font=\sffamily\fontsize{10.08}{12}\selectfont] at (325,\yy+7) {\b};
 \foreach \k in {1,...,\n}{\fill[softblue] ({347+(\k-1)*23},\yy) rectangle ++(18,14);}}
\node[anchor=west,text=softteal,font=\sffamily\bfseries\fontsize{12.5}{15}\selectfont] at (67,290.4) {Intra-block unmasking};
\node[anchor=west,text=softmuted,font=\sffamily\fontsize{10.5}{15}\selectfont] at (67,325.6) {Reveal tokens; predict the rest};
\fill[softink] (345,282) rectangle ++(18,22);
\foreach \k in {1,...,7}{\draw[softteal,fill=white] ({345+\k*23},282) rectangle ++(18,22);}
\foreach \k in {2,5,6}{\fill[softteal] ({345+\k*23},282) rectangle ++(18,22);}
\node[text=softmuted,font=\sffamily\fontsize{10.08}{12}\selectfont] at (430,325) {revealed / masked};
\node[anchor=west,text=softblue,font=\sffamily\bfseries\fontsize{12.5}{15}\selectfont] at (67,380.4) {Acceptance-weighted loss};
\node[anchor=west,text=softmuted,font=\sffamily\fontsize{10.5}{15}\selectfont] at (67,415.6) {Prefix-dependent weights};
\foreach \k/\h in {0/40,1/37,2/31,3/26,4/15,5/12,6/8,7/5}{\fill[softblue] ({345+\k*23},410) rectangle ++(18,-\h);}
\node[text=softmuted,font=\sffamily\fontsize{10.08}{12}\selectfont] at (432,428) {Position $k\;\rightarrow$};
\draw[->,softblue,line width=1.08pt] (300,447)--(300,465);
\node[text=softblue,font=\sffamily\bfseries\fontsize{13.68}{16}\selectfont] at (300,483) {Trained checkpoint {\normalfont\sffamily ($t,d_{curr}$)}};
\node[font=\fontsize{25}{28}\selectfont] at (888.5,104) {$G_{t,d,bs}=\tau_{(t,d,bs)}/T_{(t,d,bs)}$};
\node[text=softmuted,font=\sffamily\fontsize{11}{13}\selectfont] at (888.5,142) {MAT / round latency};
\node[anchor=west,text=softteal,font=\sffamily\bfseries\fontsize{14}{17}\selectfont] at (647,184) {Inference Configuration Search};
\node[anchor=west,text=softmuted,font=\sffamily\fontsize{11}{13}\selectfont] at (649,217.312) {Sweep $bs$ within one checkpoint};
\fill[softgreenbg] (649,242) rectangle ++(459,42);
\fill[softbluebg!70] (649,367.632) rectangle ++(459,42);
\foreach \y in {284,325.6,367.632}{\draw[softedge] (649,\y)--(1108,\y);}
\node[text=softteal,font=\sffamily\bfseries\fontsize{12}{14}\selectfont] at (700.5,263) {Depth};
\foreach \xx/\label in {796.5/{bs=8},885.5/{16},974.5/{24},1063.5/{32}}{\node[text=softteal] at (\xx,263) {$\label$};}
\node at (700.5,305) {$t,d_1$};
\node at (700.5,344.2) {$\cdots$};
\node[text=softblue] at (700.5,389.432) {$t,d_{curr}$};
\draw[softedge] (647,411.232)--(1106,411.232);
\draw[softblue,line width=1.08pt] (752,368.432) rectangle ++(356,42);
\foreach \xx/\b in {796.5/8,885.5/16,974.5/24,1063.5/32}{
 \node at (\xx,305) {$G_{d_1,\b}$};
 \node at (\xx,344.2) {$\cdots$};
 \node[text=softblue] at (\xx,389.432) {$G_{d,\b}$};}
\draw[->,softteal,line width=1.08pt] (877.5,420)--(877.5,440);
\node[text=softteal,font=\sffamily\bfseries\fontsize{12.5}{15}\selectfont] at (877.5,460) {Select next candidate $d$};
\node[text=softmuted,font=\sffamily\fontsize{10.5}{13}\selectfont] at (877.5,487) {Measured end-to-end speedup};
\node[font=\sffamily\bfseries\fontsize{14.4}{17}\selectfont] at (1470,108) {End-to-end speedup};
\node[text=softmuted,font=\sffamily\fontsize{10.8}{13}\selectfont] at (1470,142.08) {Qwen3-8B | temperature 0 | Conc.=1};
\foreach \name/\conf/\v/\yy/\col in {DScout/$bs=32$/4.1434917635/190/softblue,DFlash/$bs=16$/3.8696849998/267/softbar,EAGLE-3/$\mathrm{tree\ size}\ 60$/2.3896696165/344/softbar,PARD-2/$bs=15$/2.4362008203/421/softbar}{
 \node[anchor=west,font=\sffamily\bfseries\fontsize{12.96}{15}\selectfont] at (1215,\yy) {\name};
 \node[anchor=east,text=softmuted,font=\sffamily\fontsize{10.08}{12}\selectfont] at (1720,\yy) {\conf};
 \fill[\col] (1215,\yy+15) rectangle ++({\v*100},26);
 \node[anchor=west,font=\sffamily\fontsize{11.52}{13}\selectfont] at ({1223+\v*100},\yy+28) {\pgfmathprintnumber[fixed,precision=2]{\v}$\times$};}
\draw[softedge] (1215,474)--(1685,474);
\foreach \v in {0,1,2,3,4}{\draw[softmuted,line width=.6pt] ({1215+\v*100},474)--++(0,7);\node[text=softmuted,font=\sffamily\fontsize{9.36}{11}\selectfont] at ({1215+\v*100},492) {\v};}
\draw[->,softblue,line width=1.15pt,rounded corners=4pt]
 (458,483)--(592,483)--(592,388.632)--(649,388.632);
\draw[->,softblue,line width=1.05pt,rounded corners=4pt]
 (1108,460)--(1140,460)--(1140,532)--(30,532)--(30,110)--(60,110);
\node[fill=white,inner sep=3pt,text=softblue,font=\sffamily\fontsize{11.52}{13}\selectfont]
 at (590,532) {Next candidate $d$: train new drafter};
\end{tikzpicture}%
\endgroup%
}
    \caption{\textbf{Factorization-guided training and inference.}
    Training shapes position-wise supervision; inference selection balances MAT against latency. Right: six-task mean speedup for the displayed configurations from Table~\ref{tab:main}; each method uses its matched AR baseline.}
    \label{fig:dscout_pipeline}
    \vspace{-0.25in}
\end{figure}

We introduce \textbf{DScout}, which uses two complementary factorizations
to guide diffusion-drafter training and deployment-time inference optimization
(Figure~\ref{fig:dscout_pipeline}). A position-wise loss factorization
guides supervision allocation to improve MAT, while a throughput
factorization guides inference configuration selection through the
MAT--latency trade-off.

For training, a correctly predicted token contributes to MAT only if
the preceding prefix is accepted. This motivates examining not only
prediction loss, but also which positions receive supervision and how
their losses are weighted. We therefore factorize each position's loss
contribution into \textit{inclusion}, \textit{weighting}, and \textit{prediction loss}: whether the
position receives supervision, how strongly it is weighted, and how its
prediction loss is computed. Training interventions affecting different
factors can be studied separately, while those affecting the same factor
should be considered jointly to account for potential interactions.
Guided by this view, we design a recipe combining mixed-horizon training,
intra-block unmasking, and acceptance-weighted loss to align supervision
with prefix acceptance and improve MAT.
Section~\ref{sec:training-design} details the motivation and design of
each component.

For inference, we factorize throughput into MAT and latency contributions
as $G=\mathrm{MAT}/T$, where $T$ is the latency of one draft--verify round.
Guided by this factorization, we identify three broadly applicable
configuration dimensions for diffusion drafters: \textit{target-conditioning
depth}, \textit{drafter depth}, and \textit{inference block size}. We assess their joint
effects on MAT and latency to guide configuration selection.
Our cost analysis and empirical measurements motivate full target
conditioning with negligible incremental overhead, drafter depth
selection that balances MAT gains against latency, and workload-aware
inference block size selection without retraining through mixed-horizon
training. Together, these policies guide the selection of high-throughput
configurations for different models and concurrency settings.
Section~\ref{sec:deployment-design} details the analysis and selection
procedure.

Our key contributions are:
\begin{itemize}[topsep=3pt,itemsep=2pt,parsep=0pt]
    \item We formulate a position-wise loss factorization to guide
    diffusion drafter training, yielding a recipe that combines
    mixed-horizon training, intra-block unmasking, and acceptance-weighted
    loss to improve MAT without additional inference-time computation.

    \item We factorize throughput as the ratio of MAT to latency,
    guiding the joint selection of target-conditioning depth, drafter
    depth, and inference block size for high-throughput configurations
    across workloads.

    \item We validate the training recipe and inference configuration
    selection across target models, benchmarks, and concurrency settings
    through end-to-end comparisons and ablation studies.
\end{itemize}

\FloatBarrier


\section{Preliminaries}
\label{sec:preliminaries}

\subsection{Speculative Decoding}
\label{sec:prelim-acceptance}

In each speculative decoding round, a lightweight drafter proposes
$bs-1$ tokens conditioned on the verified context, where $bs$ is the
inference block size. The target verifies them in one parallel forward
pass, accepts the consecutive prefix before the first rejection, and
emits a replacement token at that position or a bonus token after full
acceptance. Appropriate acceptance and replacement rules preserve the
target model's output distribution~\citep{leviathan2023speculative,chen2023speculative}.

Let $N$ denote the number of draft tokens accepted before the first
rejection, with $0\le N\le bs-1$. The conditional acceptance probability
$a_k$ and the prefix-survival probability $S(k)$ are
\begin{align}
a_k=\Pr(N\ge k\mid N\ge k-1),
\qquad
S(k)=\Pr(N\ge k)=\prod_{j=1}^{k}a_j.
\label{eq:prelim-survival}
\end{align}
Mean accepted tokens (MAT) per round is
\begin{align}
\tau=1+\mathbb{E}[N]
=1+\sum_{k=1}^{bs-1}S(k),
\label{eq:prelim-mat}
\end{align}
where the additional one counts the target replacement or bonus token.
The product in $S(k)$ captures the prefix dependence: a draft token can
contribute to accepted length only if all preceding draft tokens are accepted.

\subsection{Block-Diffusion Drafting}
\label{sec:prelim-drafting}

Block-diffusion language models generate text by unmasking tokens within
a block using the available context. When used as speculative drafters,
they predict multiple future tokens jointly rather than generating
a proposal autoregressively. The target model then verifies the proposal
as described in Section~\ref{sec:prelim-acceptance}.

We adopt DFlash's target-conditioned block-diffusion architecture~\citep{chen2026dflash},
which predicts all masked positions in a single forward pass.
Its $bs$-position draft block contains the latest target-generated token
followed by $bs-1$ mask tokens.

To condition these predictions on the verified context, hidden states
from $t$ selected layers of the frozen target
model are concatenated and fused through a lightweight linear projection:
\begin{align}
\mathbf{h}_{\mathrm{ctx}}=W_f[h^{(\ell_1)};\ldots;h^{(\ell_t)}]+b_f,
\qquad W_f\in\mathbb{R}^{H_D\times tH_T}.
\label{eq:prelim-context-fusion}
\end{align}
Here $h^{(\ell_i)}\in\mathbb{R}^{H_T}$ represents the same context token
at target layer $\ell_i$; $H_T$ and $H_D$ are the target and drafter
hidden widths, $\mathbf{h}_{\mathrm{ctx}}\in\mathbb{R}^{H_D}$ is the
fused feature, and $b_f\in\mathbb{R}^{H_D}$ is the projection bias.
The target remains frozen. Each drafter layer uses its own projections
to map the fused features into cached keys and values reused across
rounds. Block positions attend to this context and bidirectionally
to one another.

\section{Method}
\label{sec:method}

DScout combines training recipe design with inference configuration
selection. Two complementary factorizations guide these stages:
a position-wise loss factorization for improving MAT, and a throughput
factorization for balancing MAT against draft--verify latency.

\subsection{Training Recipe Design}
\label{sec:training-design}

The prefix-survival relation in Eq.~\eqref{eq:prelim-survival} shows
that a correctly predicted token contributes to MAT only if the
preceding prefix is accepted. Improvements in token-level predictions
therefore need not yield corresponding gains in accepted length.
This motivates examining how supervision is allocated across positions
to better align training with the MAT objective.

\subsubsection{Position-Wise Loss Factorization}
\label{sec:training-factorization}


We formalize this supervision allocation through a position-wise
loss factorization.

During training, each block consists of a visible anchor token and
$B-1$ subsequent target tokens, where $B$ denotes the training horizon.
A mask pattern $m$ specifies which future tokens within the block
are masked for prediction and which are revealed as context.

For each future position $k$ relative to the anchor, let $I_k$ indicate
whether it is included in the training loss: the position must lie
within the training horizon and be masked for prediction. We factorize
its loss contribution as
\begin{align}
\mathcal{L}_k
=\underbrace{I_k}_{\text{inclusion}}
\underbrace{w(k)}_{\text{weighting}}
\underbrace{\ell(k\mid m,L)}_{\text{prediction loss}}.
\label{eq:training-factorization}
\end{align}
Here $I_k$ determines whether position $k$ receives supervision,
$w(k)$ controls how strongly that position's loss is weighted, and
$\ell(k\mid m,L)$ specifies
the prediction loss under mask pattern $m$ and loss function $L$.
The factorization accommodates different loss functions $L$.
In this work, we use token-level cross-entropy throughout and do not
explore alternative loss functions.

The training horizon $B$ and mask pattern $m$ may vary across updates.
For a fixed position $k$, $\pi(k)=\mathbb{E}_{B,m}[I_k]$ measures how
frequently it receives supervision. This exposes the position-dependent
allocation of training signal induced by horizon and mask choices.

\subsubsection{Factorization-Guided Training Strategies}
\label{sec:training-interventions}

We use the factorization to analyze training strategies in terms of
the factors they affect. For each strategy, we explain its motivation
and its effects on supervision, then examine interactions among
strategies to guide their joint design.

\begin{figure}[!htbp]
\centering
\includegraphics[width=\linewidth]{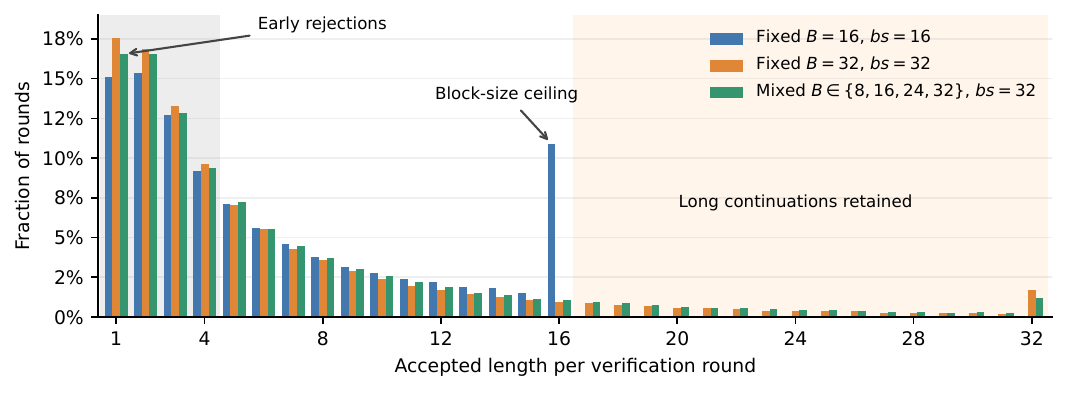}
\vspace{-0.2in}
\caption{\textbf{Mixed horizons reduce early rejections relative to fixed-32 training.} Accepted-length distributions for Qwen3-8B hidden-9t9d checkpoints trained for six epochs. Probabilities average dataset-normalized histograms equally over the six evaluation tasks. The legend specifies training horizon $B$ and inference block size $bs$; shading marks lengths 1--4 and beyond 16.}
\label{fig:horizon-acceptance-distribution}
\label{fig:mixed-horizon-distribution}
\vspace{-0.1in}
\end{figure}

\mytitle{Mixed-horizon training}
\label{sec:mixed-horizon-distribution}
Figure~\ref{fig:horizon-acceptance-distribution} shows that DFlash,
with training and inference block sizes both set to 16, accepts the
entire draft proposal in 10.9\% of verification rounds, suggesting
potential for longer continuations. Since extending inference beyond
the training horizon can reduce acceptance length~\citep{chen2026dflash},
we increase both training and inference block sizes to 32. As shown in
Figure~\ref{fig:horizon-acceptance-distribution}, this extends accepted
lengths beyond 16 tokens, but also increases the proportion of rounds
accepting only 1--4 tokens, indicating more early rejections.

This trade-off motivates mixed-horizon training, in which we use
$B\in\{8,16,24,32\}$ with equal frequency. By varying the inclusion
frequency $\pi(k)=\mathbb{E}[I_k]$, this strategy emphasizes early
positions while retaining long-horizon coverage. Varying $B$ also
changes the block context and thus the prediction-loss term
$\ell(k\mid m,L)$. As shown in
Figure~\ref{fig:horizon-acceptance-distribution}, mixed-horizon training
reduces early rejections relative to fixed-32 training while retaining
continuations beyond 16 tokens at the same inference block size.

\mytitle{Intra-block unmasking}
Random masking patterns provide data augmentation for diffusion language
models~\citep{ni2025superdata}. Revealing different subsets of tokens
exposes the model to diverse prediction contexts beyond fully masked
blocks. We apply intra-block unmasking by mixing fully masked and
partially revealed training blocks.

During training, we retain fully masked future blocks with a fixed
probability. For the remaining blocks, we sample a masking ratio from
a Beta distribution and randomly reveal the complementary subset of
future positions. Revealed positions contain
the corresponding target tokens and provide context without
contributing to the prediction loss. Unmasking changes $I_k$ by
excluding revealed positions from the loss, and $\ell(k\mid m,L)$
by providing context for the remaining predictions.

\mytitle{Acceptance-weighted loss}
Position-dependent loss decay assigns larger weights to positions
near the beginning of each draft block~\citep{chen2026dflash}, but assigns the same weight to
a given position across training examples, regardless of how likely
the preceding draft prefix is to be accepted. We introduce
acceptance-weighted loss (AWL), which dynamically weights each position's
loss based on the likelihood that verification reaches it.
AWL acts on the weighting factor $w(k)$ in
Eq.~\eqref{eq:training-factorization}, replacing fixed position-dependent
weights with prefix-dependent acceptance weights.

Let $Q_j$ denote the drafter's output distribution at position $j$
given the training block under mask pattern $m$ and the target
context features, and $P_j$ the target distribution conditioned on
its own generated prefix. For the draft prediction
$\hat y_j=\arg\max_v Q_j(v)$, we define
\begin{align*}
\alpha_j=\min\!\left(1,\frac{P_j(\hat y_j)}{Q_j(\hat y_j)}\right),
\qquad
w_k^{\mathrm{AWL}}
=\operatorname{sg}\!\left[\prod_{j<k}\alpha_j\right],
\end{align*}
where $\operatorname{sg}$ denotes stop-gradient and $w_1^{\mathrm{AWL}}=1$.
The target probability $P_j(\hat y_j)$ is evaluated under the
target-generated prefix, while $Q_j(\hat y_j)$ comes from the drafter's
prediction distribution. Both probabilities refer to the same draft
token $\hat y_j$. We compute $\alpha_j$ using the standard
speculative-sampling acceptance rule~\citep{leviathan2023speculative,chen2023speculative}.

The cumulative product of these acceptance scores reflects the
sequential nature of verification: the token at position $k$ can be
accepted only if all preceding draft tokens are accepted. An earlier
prediction with a low acceptance score therefore reduces the weights
of subsequent positions. Unlike fixed position decay, AWL dynamically
allocates supervision according to the acceptance scores of the
drafter's predictions, linking the training weights directly to
the verification rule.

We optimize the weighted mean cross-entropy over supervised positions:
\begin{align*}
\mathcal{L}_{\mathrm{train}}
=\frac{\sum_{r,k} I_{r,k}w_{r,k}^{\mathrm{AWL}}
\operatorname{CE}(Q_{r,k},y_{r,k}^{\star})}
{\sum_{r,k} I_{r,k}w_{r,k}^{\mathrm{AWL}}+\epsilon},
\end{align*}
where $r$ indexes training blocks in the minibatch, $y_{r,k}^{\star}$
is the corresponding target-generated token, and $\epsilon=10^{-6}$
ensures numerical stability.

In this objective, prefix acceptance weights scale each position's
contribution to the gradient. Early in training, the drafter's limited
predictive ability yields low acceptance scores, suppressing
later-position gradients and concentrating supervision near the start
of each block. As the drafter improves its predictions at earlier
positions, the cumulative acceptance weights of later positions
increase, progressively extending supervision to longer prefixes.
This progression aligns optimization with the verification process,
suppressing gradient contributions from later positions when their
preceding prefixes are unlikely to be accepted.


\mytitle{Joint design}
\label{sec:training-composition}
 The factorization guides not only the design of individual strategies,
but also their combination. Mixed-horizon training and intra-block
unmasking share the inclusion and prediction-loss factors, motivating
joint evaluation. Although each improves MAT alone, combining
intra-block unmasking with mixed horizons $\{4,8,16,32\}$ degrades
performance (Table~\ref{tab:horizon-unmasking}). Unmasking short blocks
may leave too few masked tokens, providing limited learning signal.
We therefore use $\{8,16,24,32\}$ to reduce short-block exposure.
The combination then outperforms either strategy alone, though its
gains are not additive, consistent with overlapping effects on shared factors.
AWL targets the weighting factor $w(k)$ and is designed separately
before being combined with the jointly designed horizon and masking
strategies. The cumulative ablation in \cref{fig:training-components}
shows their incremental contributions and combined gains.

\begin{table}[!htbp]
\centering
\begin{minipage}[t]{0.49\linewidth}
\centering
\captionsetup{font=small,skip=4pt}
\caption{Mean MAT with and without intra-block unmasking ($bs=32$).}
\vspace{-0.05in}

\label{tab:horizon-unmasking}
\footnotesize
\setlength{\tabcolsep}{2pt}
\renewcommand{\arraystretch}{1.05}
\begin{tabular}{@{}lrr@{}}
\toprule
\shortstack[l]{Training horizon} & \shortstack{w/o unmasking} & \shortstack{w/ unmasking} \\
\midrule
Fixed $32$ & 5.942 & 6.248 \\
Mixed $\{4,8,16,32\}$ & 6.228 & 6.197 \\
Mixed $\{8,16,24,32\}$ & 6.208 & \textbf{6.263} \\
\bottomrule
\end{tabular}
\end{minipage}
\hfill
\begin{minipage}[t]{0.49\linewidth}
\centering
\captionsetup{font=small,skip=4pt}
\caption{MAT across training horizons and inference block sizes.
Settings in \cref{sec:mechanism-details}.}
\vspace{-0.05in}

\label{tab:horizon}
\footnotesize
\setlength{\tabcolsep}{3pt}
\renewcommand{\arraystretch}{1.05}
\begin{tabular}{@{}lrrrr@{}}
\toprule
\shortstack[l]{Training horizon} & $bs=8$ & $16$ & $24$ & $32$ \\
\midrule
Fixed 16 & 4.61 & \textbf{6.01} & 5.87 & 5.71 \\
Fixed 32 & 4.31 & 5.29 & 5.65 & 6.18 \\
Mixed $\{8,16,24,32\}$ & \textbf{4.85} & 5.95 & \textbf{6.24} & \textbf{6.38} \\
\bottomrule
\end{tabular}
\end{minipage}
\end{table}

\subsection{Inference Configuration Selection}
\label{sec:deployment-design}

The training recipe in Section~\ref{sec:training-design} improves MAT
without introducing additional inference-time computation.
For inference, we factorize throughput into MAT and latency
contributions, highlighting the need to jointly assess their changes.
This view guides our analysis of three configuration dimensions
in diffusion drafters: target-conditioning depth $t$, drafter depth $d$,
and inference block size $bs$. Our configuration-selection objective is
\begin{align*}
(t^\star,d^\star,bs^\star)
=\arg\max_{t,d,bs}\frac{\tau(t,d,bs)}{T(t,d,bs)},
\end{align*}
where $\tau$ is MAT (Eq.~\eqref{eq:prelim-mat}), $T$ is the latency of
one draft--verify round~\citep{liu2024goodput}, and the starred variables
denote a throughput-maximizing configuration. We analyze these three dimensions to guide inference
configuration selection.

\mytitle{Full target conditioning}
\begin{figure}[!t]
\centering
\scalebox{1}[0.85]{%
  \includegraphics[width=0.9\linewidth]{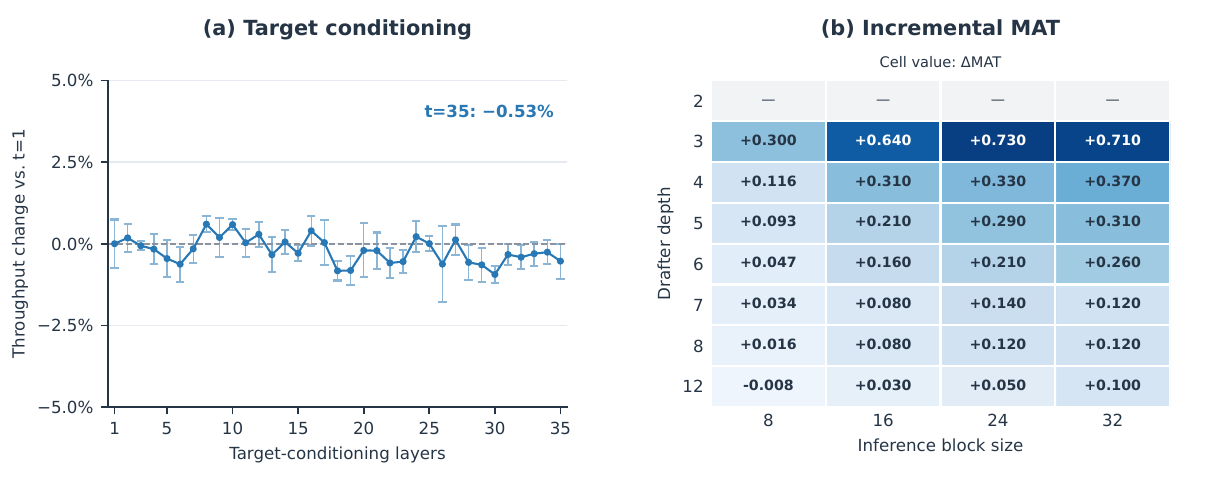}%
}
\vspace{-0.2in}
\caption{\textbf{Configuration design on Qwen3-8B.} (a) Throughput change with
target-conditioning depth at fixed acceptance. (b) Marginal MAT gains
with drafter depth; the last row spans depths 8--12. Protocols are given
in Appendix~\ref{sec:mechanism-details} and~\ref{sec:conditioning-latency}.}
\label{fig:conditioning-depth}\label{fig:conditioning-cost}
\vspace{-0.2in}
\end{figure}
Since the target model already computes hidden features at all layers,
increasing $t$ only enlarges the input to the fusion projection in
Eq.~\eqref{eq:prelim-context-fusion}, without changing subsequent
drafter computation. Our FLOPs analysis shows that, for Qwen3-8B at
$bs=16$ and MAT $\approx7$, increasing $t$ from 9 to 35 adds only
2.75\% of the target projection and MLP FLOPs
(Appendix~\ref{sec:conditioning-arithmetic}). Including
attention matrix products and drafter computation in the total would
further reduce this fraction.
Figure~\ref{fig:conditioning-depth}(a) complements this analysis with
controlled profiling at a fixed MAT of 16: throughput varies
by less than 1\% across the tested conditioning depths, with a 0.53\%
decrease from $t=1$ to $t=35$.
This small measured cost motivates using all available target features
under the tested execution conditions.

\mytitle{Unimodality-guided drafter depth selection}
Each candidate drafter depth requires training a separate drafter,
making an exhaustive depth sweep costly. We therefore seek to locate
the throughput-maximizing depth efficiently.
Our depth study reveals diminishing returns in MAT: additional drafter
layers yield progressively smaller acceptance gains
(Figure~\ref{fig:conditioning-depth}(b)). In contrast, each additional
layer introduces the same computation. At a fixed inference block
size, round latency grows approximately linearly with depth, as
supported by controlled profiling
(Appendix~\ref{sec:depth-latency-profile}).
Together, these observations suggest a unimodal throughput curve
under the MAT--latency factorization.

To formalize this, fix target conditioning and inference block size,
and let $\tau_d$ and $T_d$ denote MAT and round latency at depth $d$,
with throughput $G_d=\tau_d/T_d$ and MAT increment
$\Delta_d=\tau_{d+1}-\tau_d$. Assume diminishing MAT increments,
$\Delta_{d+1}\le\Delta_d$, and linear latency,
$T_d=T_{\mathrm{fix}}+d\,T_{\mathrm{layer}}$, where $T_{\mathrm{fix}}$
comprises target verification and the depth-independent parts of
drafting, and $T_{\mathrm{layer}}>0$ is the latency added by each
drafter layer. Adding one layer then changes throughput by
\begin{align*}
G_{d+1}-G_d
=\frac{R_d}{T_d(T_d+T_{\mathrm{layer}})},
\qquad
R_d=\Delta_d T_d-T_{\mathrm{layer}}\tau_d.
\end{align*}
Since $R_{d+1}-R_d=(\Delta_{d+1}-\Delta_d)T_{d+1}\le0$, once an
additional layer no longer increases throughput, further layers cannot
increase it either: $G_d$ is unimodal in $d$
(Appendix~\ref{sec:depth-unimodality}). Unimodality thus allows the throughput peak at each fixed block size
to be located efficiently, e.g., by binary search.

Figure~\ref{fig:depth} shows the corresponding empirical
trend: throughput rises and then falls with drafter depth, with small
local fluctuations. The highest measured throughput occurs at $d=6$
for $bs=16,24,32$, and at $d=4$ for $bs=8$.

\mytitle{Block-size selection without retraining}
Mismatch between training and inference block sizes can reduce MAT,
making configuration search costly if a separate drafter is trained
for each candidate inference block size. Mixed-horizon training
(Section~\ref{sec:mixed-horizon-distribution}) addresses this limitation.
Table~\ref{tab:horizon} shows that a drafter trained with mixed horizons
achieves higher MAT than the fixed-32 model at every tested inference
block size and remains close to the fixed-16 model at $bs=16$.
For each candidate drafter depth, the same mixed-horizon checkpoint
supports evaluation across candidate inference block sizes, enabling
joint selection of depth and block size without retraining for each
block size.

\section{Experimental Evaluation}\label{sec:experiments}\label{sec:exp}\label{sec:experimental-setup}
We evaluate DScout from two complementary perspectives: how much it accelerates inference, and how its training recipe
and inference configuration selection contribute to these improvements. 

\mytitle{Benchmarks and baselines} We evaluate on GSM8K~\citep{cobbe2021gsm8k} and MATH-500~\citep{hendrycks2021math,lightman2023verify} (Math), HumanEval~\citep{chen2021humaneval}, MBPP~\citep{austin2021mbpp}, and LiveCodeBench~\citep{jain2024livecodebench} (Code), and MT-Bench~\citep{zheng2023mtbench} (Chat).
We compare with EAGLE-3~\citep{li2025eagle3}, DFlash~\citep{chen2026dflash}, and PARD-2~\citep{an2026pard2}. Each method uses an autoregressive (AR) baseline matched in target model, hardware, runtime, sampling, and concurrency.

\mytitle{Training and implementation} We disable thinking mode for Qwen3-4B and Qwen3-8B. We train DScout, DFlash, EAGLE-3, and PARD-2 on the same 1.4M Open-PerfectBlend~\citep{labonne2024openperfectblend} examples with target-generated responses, following each baseline's official training recipe. DScout follows DFlash's parameters~\citep{chen2026dflash}, except Qwen3-8B uses our selected 35t6d configuration (Section~\ref{sec:inference-selection}). Training GPUs have 80 GB HBM and 3.35 TB/s bandwidth; main and serving benchmark GPUs have 192 GB HBM3 and 8 TB/s bandwidth. DScout, DFlash, and EAGLE-3 use SGLang~\citep{zheng2024sglang} with FlashInfer~\citep{ye2025flashinfer}; PARD-2 uses its official Transformers runtime with a separate matched AR baseline. Training settings and study protocols appear in \cref{tab:dscout-config,sec:eval-details}.

\mytitle{Metrics and configuration selection} All evaluations are repeated five times. Speedup is the ratio of mean output throughput to matched mean AR throughput, computed per dataset and then averaged equally across datasets. Study-specific aggregation procedures are detailed in Appendix~\ref{sec:eval-details}. MAT ($\tau$) is defined in Eq.~\eqref{eq:prelim-mat}. At concurrency above one, throughput measures aggregate service output. We report both $bs=16$ and $bs=32$ for DScout.

\subsection{End-to-End Performance.}\label{sec:end-to-end}
\mytitle{Benchmark performance} DScout improves mean speedup over DFlash on both Qwen3 targets and at both temperatures, reaching up to $4.14\times$ over AR and about 9\% over DFlash (Table~\ref{tab:main}). Longer proposals help when their additional accepted tokens offset the extra draft--verify cost. On Qwen3-8B at temperature 0, this trade-off favors $bs=32$ on the Math tasks, but $bs=16$ on MBPP, LiveCodeBench, and MT-Bench, where acceptance gains are smaller. The preferred block size also changes with temperature on Qwen3-4B. Thus, drafting quality varies by workload, and higher MAT alone does not determine the fastest configuration.
\begin{table}[!t]
\centering
\setlength{\tabcolsep}{1.4pt}
\caption{\textbf{Benchmark speedup and MAT on Qwen3 models.} All results use concurrency $c=1$. Parentheses denote EAGLE-3 tree size or inference block size; $\dagger$ marks the separate PARD-2 runtime.}
\label{tab:main}

\resizebox{\linewidth}{!}{%
\begin{tabular}{@{}llcccccccccccccc@{}}
\toprule
\multirow{2}{*}{Model} & \multirow{2}{*}{Method} & \multicolumn{4}{c}{\textsc{Math}} & \multicolumn{6}{c}{\textsc{Code}} & \multicolumn{2}{c}{\textsc{Chat}} & \multicolumn{2}{c}{\multirow{2}{*}{Avg.}} \\
\cmidrule(lr){3-6}\cmidrule(lr){7-12}\cmidrule(lr){13-14}
& & \multicolumn{2}{c}{GSM8K} & \multicolumn{2}{c}{MATH-500} & \multicolumn{2}{c}{HumanEval} & \multicolumn{2}{c}{MBPP} & \multicolumn{2}{c}{LCB} & \multicolumn{2}{c}{MT-Bench} & & \\
\midrule
\multicolumn{2}{l}{\textbf{Temperature = 0}} & Speedup & $\tau$ & Speedup & $\tau$ & Speedup & $\tau$ & Speedup & $\tau$ & Speedup & $\tau$ & Speedup & $\tau$ & Speedup & $\tau$ \\
\midrule
\multirow{6}{*}{Q3-4B} & EAGLE-3 \textcolor{gray}{(16)} & $2.07\times$ & $6.49$ & $2.07\times$ & $6.21$ & $1.88\times$ & $5.73$ & $1.83\times$ & $5.65$ & $1.75\times$ & $5.21$ & $1.34\times$ & $4.01$ & $1.82\times$ & $5.55$ \\
 & EAGLE-3 \textcolor{gray}{(60)} & $2.12\times$ & $7.12$ & $2.16\times$ & $6.95$ & $1.99\times$ & $6.51$ & $1.95\times$ & $6.46$ & $1.87\times$ & $6.00$ & $1.46\times$ & $4.70$ & $1.92\times$ & $6.29$ \\
 & DFlash \textcolor{gray}{(16)} & $4.51\times$ & $7.99$ & $4.43\times$ & $7.54$ & $3.48\times$ & $5.98$ & $3.39\times$ & $5.87$ & $3.15\times$ & $5.33$ & $2.14\times$ & $3.62$ & $3.52\times$ & $6.06$ \\
 & PARD-2 \textcolor{gray}{(15)}$^\dagger$ & $3.46\times$ & $8.36$ & $3.49\times$ & $8.11$ & $2.77\times$ & $6.47$ & $2.27\times$ & $6.02$ & $1.85\times$ & $5.21$ & $1.52\times$ & $3.78$ & $2.56\times$ & $6.32$ \\
 & \textbf{DScout} \textcolor{gray}{(16)} & $4.75\times$ & $8.62$ & $4.72\times$ & $8.20$ & $\mathbf{3.76\times}$ & $6.60$ & $\mathbf{3.61\times}$ & $6.40$ & $\mathbf{3.42\times}$ & $5.91$ & $\mathbf{2.28\times}$ & $3.94$ & $3.76\times$ & $6.61$ \\
 & \textbf{DScout} \textcolor{gray}{(32)} & $\mathbf{5.19\times}$ & $10.06$ & $\mathbf{4.96\times}$ & $9.16$ & $3.75\times$ & $6.99$ & $3.57\times$ & $6.72$ & $3.33\times$ & $6.11$ & $2.23\times$ & $4.09$ & $\mathbf{3.84\times}$ & $7.19$ \\
\midrule
\multirow{6}{*}{Q3-8B} & EAGLE-3 \textcolor{gray}{(16)} & $2.58\times$ & $6.87$ & $2.66\times$ & $6.76$ & $2.38\times$ & $6.15$ & $2.26\times$ & $5.93$ & $2.28\times$ & $5.79$ & $1.77\times$ & $4.51$ & $2.32\times$ & $6.00$ \\
 & EAGLE-3 \textcolor{gray}{(60)} & $2.56\times$ & $7.39$ & $2.67\times$ & $7.38$ & $2.43\times$ & $6.84$ & $2.38\times$ & $6.78$ & $2.40\times$ & $6.63$ & $1.90\times$ & $5.27$ & $2.39\times$ & $6.71$ \\
 & DFlash \textcolor{gray}{(16)} & $5.01\times$ & $8.19$ & $4.86\times$ & $7.67$ & $3.87\times$ & $6.15$ & $3.55\times$ & $5.67$ & $3.56\times$ & $5.60$ & $2.36\times$ & $3.71$ & $3.87\times$ & $6.17$ \\
 & PARD-2 \textcolor{gray}{(15)}$^\dagger$ & $3.61\times$ & $8.53$ & $1.75\times$ & $7.99$ & $2.77\times$ & $6.49$ & $2.77\times$ & $5.81$ & $1.96\times$ & $5.46$ & $1.75\times$ & $3.68$ & $2.44\times$ & $6.33$ \\
 & \textbf{DScout} \textcolor{gray}{(16)} & $5.17\times$ & $8.71$ & $5.01\times$ & $8.15$ & $4.03\times$ & $6.61$ & $\mathbf{3.61\times}$ & $5.97$ & $\mathbf{3.71\times}$ & $6.02$ & $\mathbf{2.40\times}$ & $3.89$ & $3.99\times$ & $6.56$ \\
 & \textbf{DScout} \textcolor{gray}{(32)} & $\mathbf{5.65\times}$ & $10.08$ & $\mathbf{5.44\times}$ & $9.29$ & $\mathbf{4.25\times}$ & $7.30$ & $3.56\times$ & $6.15$ & $3.62\times$ & $6.15$ & $2.34\times$ & $3.96$ & $\mathbf{4.14\times}$ & $7.16$ \\
\midrule
\multicolumn{2}{l}{\textbf{Temperature = 1}} & Speedup & $\tau$ & Speedup & $\tau$ & Speedup & $\tau$ & Speedup & $\tau$ & Speedup & $\tau$ & Speedup & $\tau$ & Speedup & $\tau$ \\
\midrule
\multirow{5}{*}{Q3-4B} & EAGLE-3 \textcolor{gray}{(16)} & $1.99\times$ & $6.09$ & $1.82\times$ & $5.36$ & $1.80\times$ & $5.37$ & $1.74\times$ & $5.26$ & $1.54\times$ & $4.51$ & $1.24\times$ & $3.65$ & $1.69\times$ & $5.04$ \\
 & EAGLE-3 \textcolor{gray}{(60)} & $2.01\times$ & $6.84$ & $1.89\times$ & $6.18$ & $1.86\times$ & $6.18$ & $1.81\times$ & $6.13$ & $1.64\times$ & $5.36$ & $1.32\times$ & $4.33$ & $1.75\times$ & $5.84$ \\
 & DFlash \textcolor{gray}{(16)} & $4.11\times$ & $7.05$ & $3.38\times$ & $5.57$ & $3.19\times$ & $5.31$ & $2.98\times$ & $5.01$ & $2.53\times$ & $4.15$ & $1.98\times$ & $3.25$ & $3.03\times$ & $5.06$ \\
 & \textbf{DScout} \textcolor{gray}{(16)} & $4.15\times$ & $7.21$ & $\mathbf{3.58\times}$ & $5.99$ & $\mathbf{3.45\times}$ & $5.86$ & $\mathbf{3.26\times}$ & $5.59$ & $\mathbf{2.74\times}$ & $4.57$ & $\mathbf{2.10\times}$ & $3.50$ & $\mathbf{3.21\times}$ & $5.45$ \\
 & \textbf{DScout} \textcolor{gray}{(32)} & $\mathbf{4.46\times}$ & $8.30$ & $3.56\times$ & $6.34$ & $3.39\times$ & $6.10$ & $3.15\times$ & $5.73$ & $2.62\times$ & $4.64$ & $2.00\times$ & $3.54$ & $3.20\times$ & $5.77$ \\
\midrule
\multirow{5}{*}{Q3-8B} & EAGLE-3 \textcolor{gray}{(16)} & $2.34\times$ & $6.35$ & $2.20\times$ & $5.75$ & $2.07\times$ & $5.47$ & $2.01\times$ & $5.40$ & $1.85\times$ & $4.82$ & $1.49\times$ & $3.90$ & $1.99\times$ & $5.28$ \\
 & EAGLE-3 \textcolor{gray}{(60)} & $2.40\times$ & $7.06$ & $2.31\times$ & $6.53$ & $2.19\times$ & $6.27$ & $2.16\times$ & $6.29$ & $2.02\times$ & $5.69$ & $1.64\times$ & $4.65$ & $2.12\times$ & $6.08$ \\
 & DFlash \textcolor{gray}{(16)} & $4.08\times$ & $6.67$ & $3.44\times$ & $5.47$ & $3.13\times$ & $5.01$ & $2.99\times$ & $4.82$ & $2.54\times$ & $4.03$ & $\mathbf{1.94\times}$ & $3.08$ & $3.02\times$ & $4.85$ \\
 & \textbf{DScout} \textcolor{gray}{(16)} & $4.13\times$ & $7.11$ & $3.49\times$ & $5.81$ & $3.23\times$ & $5.44$ & $\mathbf{3.02\times}$ & $5.11$ & $\mathbf{2.60\times}$ & $4.33$ & $1.92\times$ & $3.20$ & $3.06\times$ & $5.17$ \\
 & \textbf{DScout} \textcolor{gray}{(32)} & $\mathbf{4.41\times}$ & $7.84$ & $\mathbf{3.61\times}$ & $6.21$ & $\mathbf{3.29\times}$ & $5.70$ & $2.99\times$ & $5.22$ & $2.54\times$ & $4.36$ & $1.90\times$ & $3.26$ & $\mathbf{3.12\times}$ & $5.43$ \\
\bottomrule
\end{tabular}
}
\end{table}


\mytitle{Serving across models and concurrency} Table~\ref{tab:serving} extends the comparison to Llama-3.1-8B~\citep{grattafiori2024llama3} and Qwen3-Coder-30B-A3B~\citep{qwen2025coder30b}, with concurrency $c\in\{1,8,16,32\}$. DScout (16) uses $bs=16$, while DScout (best) selects one shared block size per target and concurrency by mean speedup across the six datasets; candidate sizes and selection details are given in Appendix~\ref{sec:selection-details}. DScout (best) achieves $2.45$--$4.14\times$ mean speedup over AR and exceeds native-block-size DFlash in all tested settings. At fixed checkpoints, block-size selection provides up to 9.75\% additional mean speedup over DScout (16), quantifying the benefit of adapting inference configuration to the serving load without retraining. DScout (16) already improves on DFlash except for Llama-3.1-8B at $c=32$, where the selection gain is largest. There, selecting a shorter block recovers the advantage despite reducing MAT; Qwen3-8B at the same concurrency also becomes faster with lower MAT. Fewer accepted tokens per round can therefore yield higher throughput when rounds finish sufficiently faster. This explains why block-size selection must account for both target model and serving load.
\begin{table}[!ht]
\centering
\ifarxivversion\else
\vspace{-0.1in}
\fi
\fontsize{8}{9.6}\selectfont
\setlength{\tabcolsep}{1.4pt}
\caption{\textbf{Serving speedup and MAT across target models and concurrency levels} at temp = 0.}
\ifarxivversion\else
\vspace{-0.1in}
\fi
\label{tab:serving}
\ifarxivversion
\def\servingtablewidth{\linewidth}
\else
\def\servingtablewidth{0.8\linewidth}
\fi
\resizebox{\servingtablewidth}{!}{%
\begin{tabular}{@{}llcccccccc@{}}
\toprule
\multirow{2}{*}{Model} & \multirow{2}{*}{Method} & \multicolumn{2}{c}{Con.1} & \multicolumn{2}{c}{Con.8} & \multicolumn{2}{c}{Con.16} & \multicolumn{2}{c}{Con.32} \\
\cmidrule(lr){3-4}\cmidrule(lr){5-6}\cmidrule(lr){7-8}\cmidrule(lr){9-10}
 & & Speedup & $\tau$ & Speedup & $\tau$ & Speedup & $\tau$ & Speedup & $\tau$ \\
\midrule
\multirow{3}{*}{Qwen3-4B} & DFlash (16) & $3.51\times$ & $6.06$ & $2.87\times$ & $6.06$ & $2.57\times$ & $6.08$ & $2.30\times$ & $6.06$ \\
 & DScout (16) & $3.75\times$ & $6.61$ & $\mathbf{3.08\times}$ & $6.59$ & $\mathbf{2.70\times}$ & $6.60$ & $\mathbf{2.45\times}$ & $6.63$ \\
 & DScout (best) & $\mathbf{3.83\times}$ & $7.19$ & $\mathbf{3.08\times}$ & $6.59$ & $\mathbf{2.70\times}$ & $6.60$ & $\mathbf{2.45\times}$ & $6.63$ \\
\midrule
\multirow{3}{*}{Qwen3-8B} & DFlash (16) & $3.87\times$ & $6.17$ & $3.16\times$ & $6.13$ & $2.74\times$ & $6.16$ & $2.38\times$ & $6.17$ \\
 & DScout (16) & $3.99\times$ & $6.56$ & $\mathbf{3.21\times}$ & $6.59$ & $\mathbf{2.79\times}$ & $6.59$ & $2.42\times$ & $6.58$ \\
 & DScout (best) & $\mathbf{4.14\times}$ & $7.16$ & $\mathbf{3.21\times}$ & $6.59$ & $\mathbf{2.79\times}$ & $6.59$ & $\mathbf{2.46\times}$ & $6.07$ \\
\midrule
\multirow{3}{*}{\shortstack[l]{Qwen3-Coder-\\30B-A3B-Instruct}} & DFlash (16) & $2.58\times$ & $6.41$ & $2.59\times$ & $6.41$ & $2.76\times$ & $6.43$ & $2.91\times$ & $6.43$ \\
 & DScout (16) & $2.65\times$ & $6.62$ & $2.62\times$ & $6.58$ & $2.82\times$ & $6.63$ & $2.95\times$ & $6.67$ \\
 & DScout (best) & $\mathbf{2.70\times}$ & $7.05$ & $\mathbf{2.68\times}$ & $7.10$ & $\mathbf{2.88\times}$ & $6.46$ & $\mathbf{2.98\times}$ & $6.10$ \\
\midrule
\multirow{3}{*}{Llama-3.1-8B-Instruct} & DFlash (10) & $3.37\times$ & $5.50$ & $2.99\times$ & $5.47$ & $2.80\times$ & $5.47$ & $2.52\times$ & $5.48$ \\
 & DScout (16) & $3.75\times$ & $6.20$ & $\mathbf{3.17\times}$ & $6.14$ & $\mathbf{2.99\times}$ & $6.17$ & $2.36\times$ & $6.18$ \\
 & DScout (best) & $\mathbf{3.76\times}$ & $6.36$ & $\mathbf{3.17\times}$ & $6.14$ & $\mathbf{2.99\times}$ & $6.17$ & $\mathbf{2.59\times}$ & $5.56$ \\
\bottomrule
\end{tabular}
}
\ifarxivversion\else
\vspace{-0.2in}
\fi
\end{table}

\subsection{Design Analysis}\label{sec:training-analysis}\label{sec:inference-selection}
\begin{figure}[!ht]
\centering
\begin{minipage}[t]{0.49\linewidth}
\centering
\vspace{0pt}
\scalebox{1}[0.85]{\resizebox{0.9\linewidth}{!}{%
\begin{tikzpicture}[x=1bp,y=1bp,font=\sffamily\fontsize{9.5}{11}\selectfont,
 every node/.style={inner sep=0pt,text=recipeink}]
\definecolor{recipeink}{HTML}{263546}
\definecolor{recipeblue}{HTML}{2878B5}
\definecolor{recipegrid}{HTML}{E6EAF0}
\path[use as bounding box] (0,0) rectangle (274,220.2);
\foreach \v in {6.0,6.5,7.0}{
 \pgfmathsetmacro{\y}{45.828+(\v-5.9)/1.25*156.78}
 \draw[recipegrid,line width=.6pt] (40.72,\y)--(261.04,\y);
 \draw[recipeink,line width=.6pt] (37.72,\y)--(40.72,\y);
 \node[anchor=east] at (34.72,\y) {\v};
}
\draw[recipeink,line width=.8pt] (40.72,202.608)--(40.72,45.828)--(261.04,45.828);
\node[rotate=90] at (8.5,124.2) {MAT $\tau$};
\draw[recipeblue,line width=1.1pt] (50.73,59.876)--(100.80,81.324)--(150.88,100.263)--(200.95,119.704)--(251.02,181.288);
\foreach \x/\y in {50.73/59.876,100.80/81.324,150.88/100.263,200.95/119.704,251.02/181.288}{
 \fill[recipeblue] (\x,\y) circle (1.7bp);
}
\foreach \x/\y/\delta in {75.77/70.60/+0.171,125.84/90.79/+0.151,175.92/109.98/+0.155,225.99/150.50/+0.491}{
 \node[text=recipeblue,font=\sffamily\bfseries\fontsize{9.5}{11}\selectfont] at (\x,\y+14) {\delta};
}
\foreach \x/\lab in {50.73/{$B=16$},100.80/{$B=32$},150.88/{+ Mixed},200.95/{+ Unmask},251.02/{+ AWL}}{
 \draw[recipeink,line width=.6pt] (\x,45.828)--(\x,42.828);
 \node[anchor=north,font=\sffamily\fontsize{8.5}{10}\selectfont] at (\x,38.8) {\lab};
}
\node at (150.88,21) {Cumulative training recipe};
\end{tikzpicture}}}
\captionsetup{font=small,skip=4pt}
\caption{\textbf{Cumulative recipe ablation (9t9d).} Labels show MAT increments from the preceding step.}
\label{fig:training-components}\label{tab:training-components}
\end{minipage}\hfill
\begin{minipage}[t]{0.49\linewidth}
\centering
\vspace{-0.2pt}
\scalebox{1}[0.85]{%
  \includegraphics[width=0.9\linewidth]{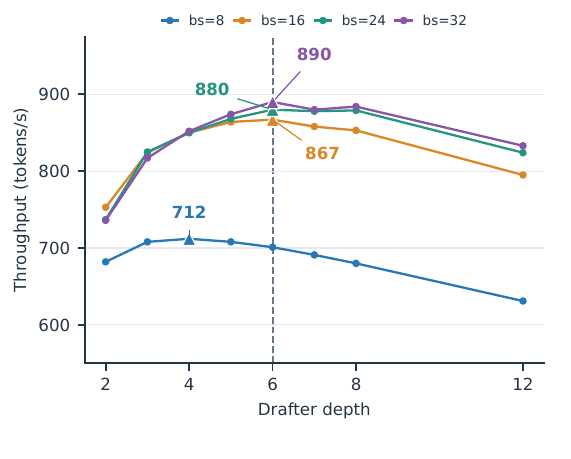}%
}
\captionsetup{font=small,skip=4pt}
\caption{\textbf{Throughput versus drafter depth} on Qwen3-8B. Markers highlight observed maxima.}
\label{fig:depth}\label{fig:depth-throughput}
\end{minipage}
\end{figure}

\mytitle{Training-recipe ablation}
Following Section~\ref{sec:training-design}, we progressively add mixed-horizon
training, intra-block unmasking, and AWL to the training recipe.
Each configuration is trained independently to produce a separate checkpoint.
At a common inference block size of $bs=16$, the cumulative training
ablation improves MAT by 16.1\% (Figure~\ref{fig:training-components}).
Every added strategy contributes, with AWL producing the largest
increment after mixed horizons and intra-block unmasking. These gains
require no additional inference-time operations; the increments are
conditional on the preceding recipe rather than independent effects.
The system-level comparison in Appendix Figure~\ref{fig:conditional}
shows higher acceptance at every position, with larger gains later in the block.

\mytitle{Drafter-depth selection}
Figure~\ref{fig:depth} shows throughput increasing and then declining
with drafter depth, with small local fluctuations. The highest measured
throughput occurs at $d=6$ for $bs=16,24,32$, but at $d=4$ for $bs=8$,
showing that the preferred depth depends on inference block size.
The unimodality result applies under its stated assumptions; local
measurement fluctuations need not satisfy it exactly.
Together with the target-conditioning analysis in
Section~\ref{sec:deployment-design}, these results motivate our 35t6d
configuration for Qwen3-8B, with inference block size selected for each
concurrency setting.

\section{Related Work}
\label{sec:related}
\mytitle{Autoregressive and parallel drafters}
Early speculative decoding methods generate proposals
autoregressively~\citep{leviathan2023speculative,chen2023speculative}.
The EAGLE family uses target-model features for drafting, with EAGLE-3
combining multi-layer features and direct token
prediction~\citep{li2024eagle,li2024eagle2,li2025eagle3}.
Other approaches reduce sequential drafting through multiple prediction
heads, as in Medusa and Hydra~\citep{cai2024medusa,ankner2024hydra},
or parallel masked-token prediction, as in PARD~\citep{an2025pard}.

\mytitle{Block-diffusion drafters}
Diffusion language models support parallel token
prediction~\citep{arriola2025block,nie2025llada} and have been explored
for speculative drafting in DiffuSpec, SpecDiff-2, and
DART~\citep{li2025diffuspec,sandler2025specdiff2,liu2026dart}.
DFlash generates proposals in a single pass using a target-conditioned
block-diffusion drafter~\citep{chen2026dflash}.
Building on this architecture, \sys studies how training supervision
and inference configuration together determine decoding efficiency.
Its two factorizations guide supervision toward accepted prefixes
and configuration selection according to the MAT--latency trade-off.

\mytitle{Serving efficiency and inference configuration}
vLLM and SGLang improve serving efficiency through memory management
and scheduling~\citep{kwon2023paged,zheng2024sglang}, while SpecInfer
supports efficient token-tree verification~\citep{miao2023specinfer}.
Complementary to these systems, \sys selects target-conditioning depth,
drafter depth, and inference block size based on their MAT--latency
trade-offs.
Mixed-horizon training further allows inference block size to be
selected for different workloads and concurrency settings without
retraining the checkpoint.

\section{Discussion and Conclusion}
\label{sec:conclusion}Our current inference selection uses a shared block size for each
target model and concurrency setting. A natural next step is to
adapt block size online as acceptance and serving load change.
Mixed-horizon checkpoints support this flexibility without retraining.
An online policy could use recent acceptance statistics and measured
round latency to adjust block size within a generation according to
the MAT--latency trade-off. Evaluating such a policy under changing
workloads, while accounting for adaptation overhead, would establish
whether it yields sustained gains over fixed configurations.

Overall, DScout provides a systematic approach to translating diffusion-drafter
prediction quality into practical decoding acceleration. For training,
a position-wise loss factorization guides strategy design and combination.
For inference, the MAT--latency factorization guides configuration
selection. The resulting training recipe and inference configurations
improve end-to-end performance across models, tasks, and serving loads.
These factorizations can also inform future diffusion-drafter research
by providing a framework for training strategy design and inference
configuration selection. 

\bibliographystyle{plainnat}
\bibliography{references}

\newpage  

\setlength{\abovedisplayskip}{10pt}
\setlength{\abovedisplayshortskip}{10pt}
\setlength{\belowdisplayskip}{10pt}
\setlength{\belowdisplayshortskip}{10pt}

\normalsize
\appendix
\raggedbottom
\part*{Supplementary Material}

{
\setlength{\parskip}{-0em}
\startcontents[sections]
\printcontents[sections]{ }{1}{}
}

\section{Evaluation Details and Supplementary Results}
\label{sec:eval-details}

\subsection{Additional Setup and Aggregation}
\label{sec:aggregation-details}

\mytitle{Evaluation protocol} We use the six evaluation tasks defined in Section~\ref{sec:experimental-setup}, with 128 prompts per task except MT-Bench (80), dataset seed 0, an 8192-token context limit, and a 2048-token generation cap. Sampling uses top-$p=1$, with top-$k=1$ at temperature 0 and no top-$k$ truncation at temperature 1; the corresponding sampling seeds are 1000 and 2000. SGLang runs use 32 warm-up prompts; PARD-2 uses one. DScout, DFlash and EAGLE-3 are trained for six epochs, and PARD-2 for two. These are complete-system comparisons with unequal training budgets. Each comparison uses its matched AR baseline; speedups across the SGLang and PARD-2 runtimes do not establish an absolute throughput ranking. Training-control studies and system benchmarks use separate platforms, and their latencies are not pooled.

\mytitle{Run variability} Point estimates follow the task-level ratio-of-means definition in Section~\ref{sec:experimental-setup}. For repeat variability, we instead compute one macro speedup per repeat by pairing candidate and AR runs with the same repeat index, then report the sample SD over five repeats. This measures variability on fixed prompts, not uncertainty over unseen tasks. Mechanism studies with different aggregation are specified below.

\subsection{Serving Configuration Grid}
\label{sec:serving-details}
\label{sec:selection-details}

We evaluate $bs\in\{8,10,12,14,16,24,32\}$ for each target
and concurrency setting.

Figure~\ref{fig:serving-grid} records every block-size candidate underlying DScout (best) in Table~\ref{tab:serving}. Each target--concurrency setting keeps its checkpoint fixed and selects one block size by maximizing equal-weight mean speedup across all six tasks. Thus, these results evaluate block-size selection at fixed $(t,d)$, not a joint optimum over all three dimensions. Selection uses the evaluation tasks themselves, without combining per-task optima. Table~\ref{tab:serving-run-sd} reports repeat variability; entries are DFlash / DScout (best).

\ifarxivversion
\begin{figure}[H]
\else
\begin{figure}[!htbp]
\fi
\centering
\ifarxivversion
\includegraphics[width=0.82\linewidth]{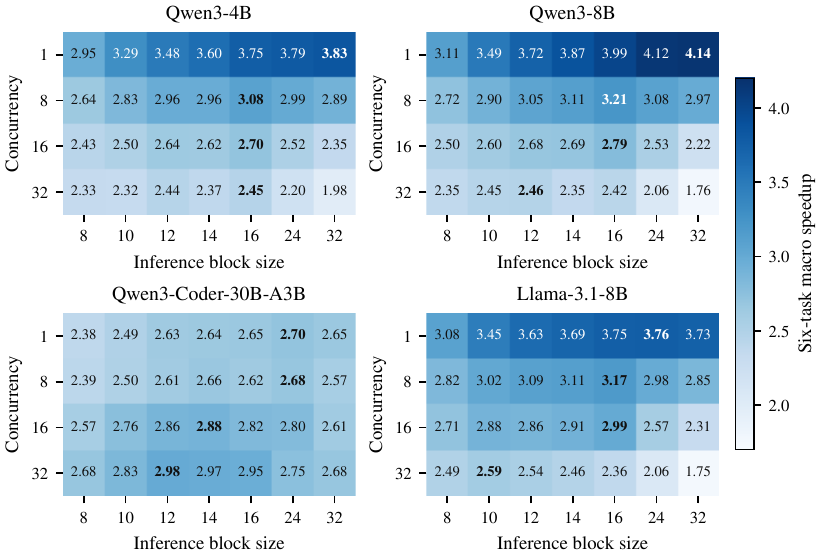}
\else
\includegraphics[width=\linewidth]{figs/factorization/serving_grid.pdf}
\fi
\caption{\textbf{Full serving grid.} Cells show six-task mean speedup; bold entries mark the selected block size in each row.}
\label{fig:serving-grid}
\end{figure}

\ifarxivversion
\begin{table}[H]
\else
\begin{table}[!htbp]
\fi
\centering\footnotesize
\setlength{\tabcolsep}{5pt}
\caption{Serving speedup: sample SD over five repeats.}
\label{tab:serving-run-sd}
\begin{tabular}{@{}lcccc@{}}
\toprule
Target & $c=1$ & $c=8$ & $c=16$ & $c=32$\\
\midrule
Qwen3-4B & .004 / .011 & .043 / .045 & .116 / .127 & .052 / .028\\
Qwen3-8B & .004 / .017 & .013 / .010 & .072 / .064 & .035 / .036\\
Qwen3-Coder-30B-A3B & .002 / .004 & .023 / .013 & .025 / .022 & .042 / .033\\
Llama-3.1-8B & .004 / .003 & .023 / .038 & .112 / .095 & .208 / .161\\
\bottomrule
\end{tabular}
\end{table}

\subsection{Acceptance and Depth Study Protocols}
\label{sec:mechanism-details}

\mytitle{Training controls} The training-horizon controls use Qwen3-8B hidden-9t9d checkpoints trained for six epochs. Fixed-16 uses positional decay $\gamma=7$; fixed-32 and mixed horizons use $\gamma=15$. Mixed horizons sample $B\in\{8,16,24,32\}$ equally. The distribution in Figure~\ref{fig:horizon-acceptance-distribution} averages dataset-normalized histograms over the same six tasks as the system benchmarks.

\mytitle{Position-wise acceptance} Figure~\ref{fig:conditional} compares final DScout (35t6d) with DFlash on Qwen3-8B. Let $S(k)$ be the probability that the first $k$ draft tokens survive verification, with $S(0)=1$. We compute conditional acceptance as $S(k)/S(k-1)$ from each dataset's accepted-length histogram in each run, then average equally over five repeats and datasets within each domain. Conditioning on surviving prefixes means these probabilities need not decrease with position. The comparison evaluates complete systems and does not isolate architecture, training-recipe, or training-budget effects.

\begin{figure}[!htbp]
\centering
\includegraphics[width=\linewidth]{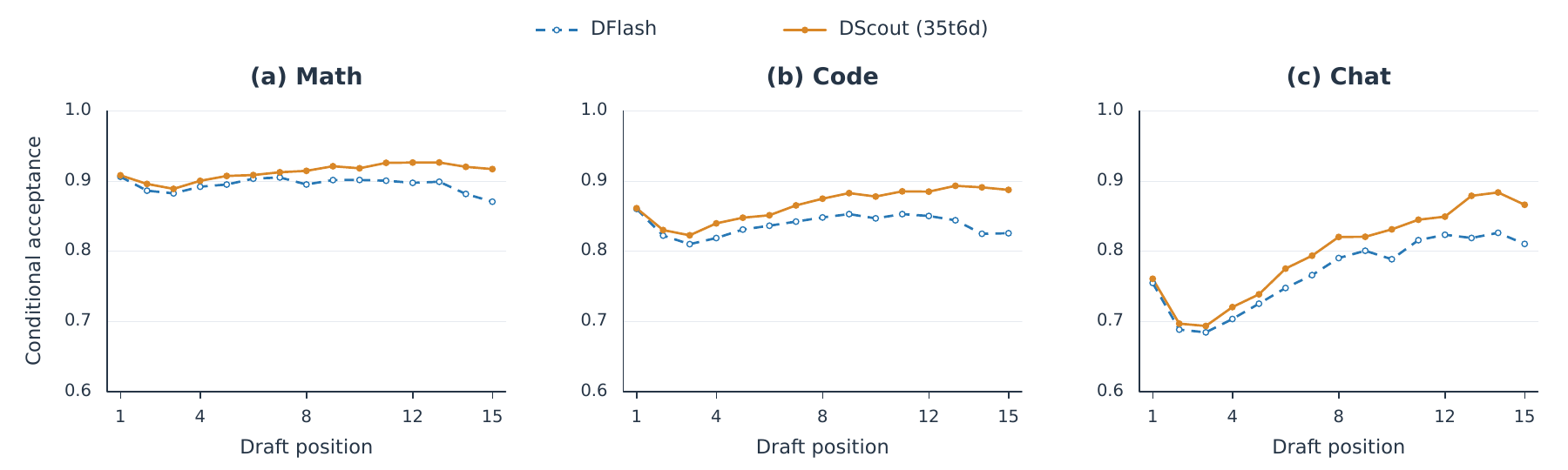}
\caption{\textbf{Position-wise conditional acceptance} of DScout (35t6d) and DFlash at $bs=16$, $c=8$, and temperature 0.}
\label{fig:conditional}
\end{figure}

\mytitle{Drafter depth} Figure~\ref{fig:conditioning-depth}(b) and Figure~\ref{fig:depth} use the Qwen3-8B 35t hidden-conditioned family trained for six epochs, with $c=1$, $d\in\{2,3,4,5,6,7,8,12\}$, and $bs\in\{8,16,24,32\}$. For each dataset, we discard the minimum and maximum of five repeats and average the middle three, then average equally across the six datasets. Figure~\ref{fig:depth} reports throughput in tokens per second (TPS); repeat-level dispersion is unavailable for these aggregates, so no error bars are shown. Highlighted maxima identify the best measured candidates, without implying statistically significant differences.

\subsection{Target-Conditioning Cost}
\label{sec:conditioning-latency}
\label{sec:conditioning-arithmetic}

\mytitle{Measured throughput} Figure~\ref{fig:conditioning-cost}(a) sweeps $t=1,\ldots,35$ at fixed $d=6$, $bs=16$, and $c=1$, with five repeats per setting. Acceptance is held fixed at 16 tokens per round to isolate execution cost from prediction quality. Let $q(t)$ be trimmed-mean throughput and $s(t)$ the across-repeat SD. Points show $100[q(t)/q(1)-1]$, and bars show $\pm100s(t)/q(1)$ percentage points. The reference mean is fixed, so the bars represent run variability rather than confidence intervals for the ratio. These measurements capture end-to-end throughput, not isolated feature-fusion latency.

\mytitle{Illustrative arithmetic cost} For concatenation followed by a linear projection, increasing conditioning from 9 to 35 layers adds
\begin{equation*}
\Delta F_{\mathrm{fusion}}=2(35-9)H^2A,
\qquad
F_{\mathrm{target,dense}}=2L_T\,bs\,(2H^2+2HK+3HI),
\end{equation*}
where $A$ is the number of newly fused context tokens and a multiply-add counts as two FLOPs. The target term counts target projection and MLP FLOPs. Here $L_T$ is the target layer count, $I$ is the MLP intermediate width, and $K$ is the total key/value projection width. We use $H_T=H_D=H$ for equal target and drafter hidden widths. Using Qwen3-8B's $L_T=36$, $H=4096$, $I=12288$, and $K=1024$,\footnote{Model dimensions: \href{https://huggingface.co/Qwen/Qwen3-8B/blob/main/config.json}{Qwen3-8B configuration}.} with $bs=16$ and $A\approx7$ corresponding to MAT near seven when only newly accepted context tokens are fused, these costs are 6.11 and 222.26 GFLOPs, respectively (2.75\%); fusing all 16 positions gives 6.28\%. This cached-context calculation excludes prefill, attention matrix products, the output head, drafter operations, and memory traffic. It illustrates arithmetic scale, not a measured fusion profile or a bound on latency; feature transfer and kernel behavior remain relevant.

\subsection{Controlled Depth--Latency Profiling}
\label{sec:depth-latency-profile}

To examine execution cost independently of prediction quality, we profile
drafter depths $d=1,\ldots,35$ with a fixed acceptance length. Let $q_d$
denote the measured throughput. Under this fixed-acceptance protocol,
relative round latency is $T_d/T_1=q_1/q_d$. Figure~\ref{fig:depth-latency-profile}
plots this ratio with an unconstrained least-squares affine fit,
$T_d/T_1\approx0.9777+0.0190d$, yielding $R^2=0.9975$.
The maximum relative deviation from the fitted line is 1.86\%.
These measurements support approximately linear latency over the profiled
depth range, complementing the MAT increments and end-to-end throughput
in Figure~\ref{fig:conditioning-depth}(b) and Figure~\ref{fig:depth}.

\begin{figure}[!htbp]
\centering
\includegraphics[width=0.7\linewidth]{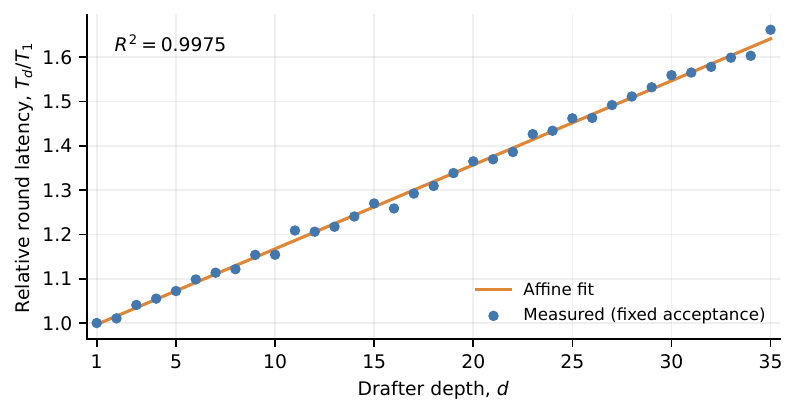}
\caption{\textbf{Depth--latency profile.} Measured relative round latency and affine fit.}
\label{fig:depth-latency-profile}
\end{figure}

\FloatBarrier

\subsection{Conditional Unimodality over Drafter Depth}
\label{sec:depth-unimodality}

Fix target conditioning, inference block size, workload, and execution
conditions. On a finite interval of consecutive drafter depths, let
$\tau_d>0$ be MAT, $T_d>0$ be mean round latency, and
$\Delta_d=\tau_{d+1}-\tau_d$. Write the per-layer latency increment as
$b_d=T_{d+1}-T_d>0$. Assume diminishing MAT increments,
$\Delta_{d+1}\le\Delta_d$, and non-decreasing latency increments,
$b_{d+1}\ge b_d$. Then throughput $G_d=\tau_d/T_d$ is unimodal
in the sense that it is non-decreasing up to some depth and
non-increasing thereafter (equivalently, quasi-concave in $d$): its increments cannot become positive after
becoming nonpositive. Plateaus and boundary optima are permitted; MAT need not be
monotonically increasing. The linear latency model in the main text
is the special case $b_d=T_{\mathrm{layer}}$.

\paragraph{Proof.}
Since $T_d>0$ and $T_{d+1}>0$,
\begin{equation*}
G_{d+1}-G_d=\frac{R_d}{T_dT_{d+1}},
\qquad
R_d=\Delta_dT_d-b_d\tau_d,
\end{equation*}
so the sign of each throughput increment is the sign of $R_d$.
Using $\tau_{d+1}=\tau_d+\Delta_d$ and $T_{d+1}=T_d+b_d$ gives
\begin{equation*}
R_{d+1}-R_d
=(\Delta_{d+1}-\Delta_d)T_{d+1}-(b_{d+1}-b_d)\tau_{d+1}\le0,
\end{equation*}
where the first term is nonpositive by diminishing MAT increments and
the second is nonpositive because latency increments do not decrease
and $\tau_{d+1}>0$. Thus $R_d$ is non-increasing, and once $R_d\le0$,
every subsequent throughput increment is nonpositive. This proves the
claim. Under linear latency, the second term vanishes, so the sign
change is driven by the diminishing MAT increments alone.

\paragraph{Relation to the measurements.}
Figure~\ref{fig:conditioning-depth}(b) motivates diminishing returns
over the consecutive measured depths 2--8. Its final row reports the
total gain from depth 8 to 12, not a one-layer increment, and does not
resolve the unmeasured intermediate depths. Figure~\ref{fig:depth} shows throughput
rising and then falling with small local fluctuations; it does not by itself
establish exact unimodality. The separate fixed-acceptance profile in
Appendix~\ref{sec:depth-latency-profile} supports approximately linear latency.
The proposition provides a search rationale under explicit assumptions,
rather than an unconditional guarantee for the measured serving curves.
It applies separately at each fixed block size: it does not establish
unimodality of the envelope $\max_{bs}G(d,bs)$ or a global optimum over
all configurations.

\section{DScout algorithm}
\label{sec:append}

\begin{algorithm}[H]
\captionsetup{justification=raggedright,singlelinecheck=false}
\caption{DScout training}
\label{alg:dscout-training}
\small
\begin{algorithmic}[1]
\Require Target-aligned dataset $\mathcal{D}$ and frozen target $p_T$
\Require Drafter $q_\theta$ and context adapter $g_\phi$

\For{each minibatch $(x,y)\sim\mathcal{D}$}
    \State Choose $B\in\{8,16,24,32\}$ with equal frequency across updates
        and valid anchors $\mathcal{A}$
    \State $(H_{\mathcal S},P)\gets p_T(x,y)$
    \hfill\textit{target frozen}
    \State $C\gets g_\phi(H_{\mathcal S})$
    \State $U\gets0$;\quad $Z\gets0$
    \hfill\textit{weighted-loss and weight accumulators}

    \For{each anchor $a\in\mathcal{A}$}
        \State $Y\gets(y_a,\ldots,y_{a+B-1})$
        \State $M\gets\mathrm{SampleMask}(B)$
        \State $\widetilde{Y}\gets\mathrm{Corrupt}(Y,M)$
        \State $Q_{1:B-1}\gets
            q_\theta(\widetilde{Y},C_a)$

        \For{$k=1,\ldots,B-1$}
            \State $\hat y_k\gets\arg\max_v Q_k(v)$
            \State $\alpha_k\gets
                \min\!\left\{
                    1,\,
                    \dfrac{P_k(\hat y_k)}
                          {Q_k(\hat y_k)}
                \right\}$
            \State $w_k\gets
                \operatorname{sg}\!\left(
                    \prod_{j<k}\alpha_j
                \right)$
        \EndFor

        \State $U\gets U+
            \displaystyle\sum_{k\in M}
            w_k\,\mathrm{CE}(Q_k,y_{a+k})$
        \State $Z\gets Z+
            \displaystyle\sum_{k\in M}w_k$
    \EndFor

    \State $\mathcal{L}_{\mathrm{train}}
        \gets U/(Z+\epsilon)$
    \State Update $\theta$ and $\phi$ using
        $\nabla\mathcal{L}_{\mathrm{train}}$
\EndFor
\end{algorithmic}
\end{algorithm}

\section{Training hyperparameters}

\begin{table}[H]
\captionsetup{justification=raggedright,singlelinecheck=false}
\caption{DScout training configurations.}
\label{tab:dscout-config}
\centering
\small
\begin{tabularx}{\linewidth}{
    @{}
    >{\raggedright\arraybackslash}p{0.36\linewidth}
    >{\raggedright\arraybackslash}X
    @{}
}
\toprule
\textbf{Configuration} & \textbf{Value} \\
\midrule

\multicolumn{2}{@{}l}{\textit{Target-aligned data}} \\
Dataset
    & Open-PerfectBlend~\citep{labonne2024openperfectblend} \\
Number of samples
    & $1.4$M (\texttt{1m4}) \\
Rollout temperature
    & $0$ \\

\addlinespace
\multicolumn{2}{@{}l}{\textit{Coverage-oriented training}} \\
Anchors per sequence
    & $512$ \\
Training horizon
    & $B\in\{8,16,24,32\}$, equal frequency \\
Full-mask probability
    & $0.4$ \\
Diffusion mask distribution
    & $\rho\sim\mathrm{Beta}(2,2)$ \\
Minimum mask fraction
    & $0.25$ \\

\addlinespace
\multicolumn{2}{@{}l}{\textit{Position weighting and normalization}} \\
Position weighting
    & AWL\\
Weight normalization
    & Weighted mean \\
Diffusion-loss normalization
    & Weighted mean \\

\addlinespace
\multicolumn{2}{@{}l}{\textit{Optimization}} \\
Training epochs
    & $6$ \\
Optimizer
    & AdamW \\
Learning rate
    & $6\times10^{-4}$ \\
Warmup ratio
    & $0.04$ \\
Gradient clipping
    & $1.0$ \\
Global batch size
    & $64$ \\

\bottomrule
\end{tabularx}
\end{table}


\end{document}